\documentclass{article}
\usepackage{spconf,amsmath,amssymb,graphicx}
\usepackage{array, booktabs}
\usepackage{cite}
\usepackage{bm}
\usepackage{hyperref}
\usepackage{algorithm, algpseudocode}
\algnewcommand{\Initialize}[1]{%
    \State \textbf{Initialize:}
    \Statex \hspace*{\algorithmicindent}\parbox[t]{.8\linewidth}{\raggedright #1}
}

\def\q{{\bm q}}
\def\x{{\bm x}}
\def\c{{\bm c}}
\def\P{{\cal P}}
\title{FORWARD ATTENTION IN SEQUENCE-TO-SEQUENCE ACOUSTIC MODELING FOR SPEECH SYNTHESIS}
\name{}

\name{Jing-Xuan Zhang, Zhen-Hua Ling, Li-Rong Dai
\thanks{This work was partially funded by the Fundamental Research Funds for the
Central Universities (Grant No. WK2350000001), and the National
Nature Science Foundation of China (Grant No. U1636201).}}
\address{National Engineering Laboratory for Speech and Language Information Processing
\\University of Science and Technology of China, Hefei, P.R.China\\
{\small \tt nosisi@mail.ustc.edu.cn, \{zhling,lrdai\}@ustc.edu.cn}}
\begin{document}
\ninept
\maketitle
\begin{abstract}
This paper proposes a forward attention method for the sequence-to-sequence acoustic modeling of speech synthesis.
This method is motivated by the nature of the monotonic alignment from phone sequences to acoustic sequences.
Only the alignment paths that satisfy the monotonic condition are taken into consideration at each decoder timestep.
The modified attention probabilities at each timestep are computed recursively using a forward algorithm.
A transition agent for forward attention is further proposed, which helps the attention mechanism to make decisions whether to move forward or stay at each decoder timestep.
Experimental results show that the proposed forward attention method achieves faster convergence speed and higher stability than the baseline attention method.
Besides, the method of forward attention with transition agent can also help improve the naturalness of synthetic speech and control the speed of synthetic speech effectively.

\end{abstract}
\section{Introduction}
\label{sec:intro}
A statistical parametric speech synthesis (SPSS) \cite{King2011,taylor2009text,zen2009statistical} system typically consists of a text analysis
frontend, an acoustic model, a duration model and a vocoder for waveform reconstruction.
The task of the acoustic model is to convert linguistic input into acoustic output.
In conventional neural-network-based acoustic modeling \cite{40837,fan2014tts,43893,watts2016hmms,7078992},
we usually align a linguistic feature sequence and
the corresponding acoustic trajectory by a hidden Markov model (HMM) at first due to the different lengths of these two feature sequences.
Then, a deep neural network (DNN) or long short-term memory (LSTM)-based \cite{hochreiter1997long} acoustic model can be built using the aligned frame-level
input-output pairs.
Besides, a separate duration model is always necessary to predict the duration of HMM states or phones at synthesis time.

On the other hand, sequence-to-sequence (seq2seq) neural networks \cite{sutskever2014sequence,cho2014learning} have been proposed recently,
which can transduce an input sequence directly into an output sequence that may have different length.
Encoder-decoder with attention is the most popular architecture to achieve seq2seq modeling at current stage.
It has been successfully applied to various tasks,
such as machine translation \cite{bahdanau2014neural,luong2015effective},
image caption generation \cite{xu2015show}
and speech recognition
\cite{chorowski2015attention, chan2016listen, jaitly2016online}.

The seq2seq modeling techniques have also been applied to speech synthesis in the last two years \cite{wang2016first,Sotelo2017,wang2017tacotron}.
To our knowledge, the first work among them \cite{wang2016first} adopted content-based attention \cite{bahdanau2014neural} to build the encoder-decoder acoustic model for speech synthesis.
The windowing technique and convolutional features \cite{chorowski2015attention} were also used to stabilize the attention alignment.
Char2Wav \cite{Sotelo2017} employed location-based attention \cite{graves2013generating}.
Tacotron \cite{wang2017tacotron} improved the network architecture of encoder and decoder, and adopted a reduction trick to help the attention moving forward without getting stuck.
There are several advantages of these seq2seq models for speech synthesis.
First, we can train acoustic models from scratch data conveniently, which helps to build end-to-end systems without explicit text analysis modules.
Second, the separate duration model is not necessary any more. To predict acoustic features with appropriate durations from a unified model may lead to better naturalness of synthetic speech.

Speech synthesis can be considered as a \emph{decompressing} process, i.e.,
one input phone should be translated into tens of acoustic frames.
Therefore, it is a challenge for the attention mechanism to
keep focus on one phone for many decoder timesteps and go forward step by step.
Current seq2seq models for speech synthesis still suffer from
the issue of instability, such as missing phones and repeating phones in the synthetic speech or even failing to generate intelligible speech.
Besides, without a separate duration model, it is difficult to control the speed of synthetic speech using seq2seq acoustic models.

Therefore, this paper proposes a forward attention method for the seq2seq acoustic modeling of speech synthesis.
This method is motivated by the nature of the monotonic alignment from phone sequences to acoustic sequences.
Only the alignment paths that satisfy the monotonic condition are taken into consideration at each decoder timestep.
The modified attention probability at each timestep can be computed recursively using a forward algorithm.
Furthermore, a transition agent for forward attention is proposed, which helps the attention mechanism to make decisions whether to move forward or stay at each decoder timestep.

Overall, the contributions of this paper are two-fold.
First, we propose a new forward attention method, 
which achieves faster convergence speed, better stability of acoustic feature generation, and higher naturalness of synthetic speech than baseline attention method.
Second, we can control the speed of synthesized speech based on the proposed forward attention
 method, which is difficult for the original content-based attention method.
\nopagebreak[4]
\section{Previous Work}
\label{sec:previous works}


A model of encoder-decoder with attention \cite{bahdanau2014neural,luong2015effective} converts an input sequence into an output target sequence with different length.
Encoders and decoders are usually recurrent neural networks (RNN).
The encoder first processes the input sequence $\bm{t} = [\bm{t}_1,\bm{t}_2,...,\bm{t}_N]$
 to produce a sequence of hidden representations
 $\bm{x}=[\bm{x}_1,\bm{x}_2,...,\bm{x}_N]$ which are
more suitable for the attention mechanism to work with.
The decoder then generates the output sequences
$\bm{o}=[\bm{o}_1,\bm{o}_2,...,\bm{o}_T]$,
conditioning on $\x$.

At each decoder timestep $t$, the attention mechanism uses an internal inference step
to perform a soft-selection over these representations \cite{kim2017structured}.
Let $\q_t$ denote the query of the output sequence at the $t$-th timestep which is usually the hidden state of the decoder RNN,
and $\pi_t \in \{1,2,...,N\}$ be a categorical latent variable that represents the selection
among hidden representatioins according to the conditional distribution $p(\pi_t|\x,\q_t)$.
The context vector derived from the input is defined as
\begin{equation}
\c_t=\sum_{n=1}^N{y_t(n)\x_{n}},\label{eq1}
\end{equation}
where $y_t(n)=p(\pi_t=n|\x,\q_t)$.
Finally, the output vector $\bm{o}_t$ can be computed conditioning on the context $\c_t$.
In the widely-used content-based attention mechanism \cite{bahdanau2014neural},
$p(\pi_t|\x,\q_t)$ is calculated as
\begin{equation}
e_{t,n}=\bm{v}^T\text{tanh}(\bm{W}\q_t + \bm{V}\x_n + \bm{b}),\end{equation}
\begin{equation}
y_t(n) =
\left.\text{exp}(e_{t,n})\middle/
\sum_{m=1}^N{\text{exp}(e_{t,m})}.\right.
\end{equation}

Some techniques have been proposed to improve the performance of original attention mechanism.
One is \emph{adding convolutional features }\cite{chorowski2015attention} 
for stabilizing the attention alignment.
In detail, $k$ filters with kernel size $l$ are employed to convolve the alignment of previous decoder timestep.
Let $\bm{F}\in \mathbb{R}^{k\times l}$ represent the convolution matrix.
Then it is used as an extra term for calculating the attention probabilities and we have
\begin{equation}
\bm{f}_t=\bm{F}*\bm{y}_{t-1},
\end{equation}
\begin{equation}
e_{t,n}=\bm{v}^T\text{tanh}(\bm{W}\q_t + \bm{V}\x_n + \bm{U}\bm{f}_{t,n} + \bm{b}),
\end{equation}
where $*$ denotes convolution and $\bm{y}_t=[y_t(1),\dots,y_t(N)]^\top$.

Another technique is \emph{windowing} \cite{chorowski2015attention}. Only a subset of
the encoding results  $\hat{\x}=[\x_{p-w}, ..., \x_{p + w}]$ are considered at each decoder timestep when using the windowing
technique. Here, $w$ is the
window width and $p$ is the middle position of the window, e.g., the mode of the alignment probability of previous decoder timestep.
This technique can not only stabilize the attention alignment but also reduce the computational complexity.

In the application of speech synthesis, the alignment path $\{\pi_1,\pi_2,...,\pi_T\}$ between $\x$ and $\bm{o}$ indicates how input linguistic features are mapped to their corresponding acoustic features.
When phone sequences are used as the input, we expect that the attention
should focus on one phone to generate context vectors for tens of acoustic frames, and then move forward to the next phone along a monotonic direction.
Therefore, we will propose a new forward attention method for the seq2seq acoustic modeling of speech synthesis in the next section.

\section{Forward Attention for Sequence-to-Sequence Modeling}
\label{sec:forward attention}


\subsection{Forward Attention}
\label{ssec:forward attention model}

\begin{figure}
  \centering
  \includegraphics[scale=0.6]{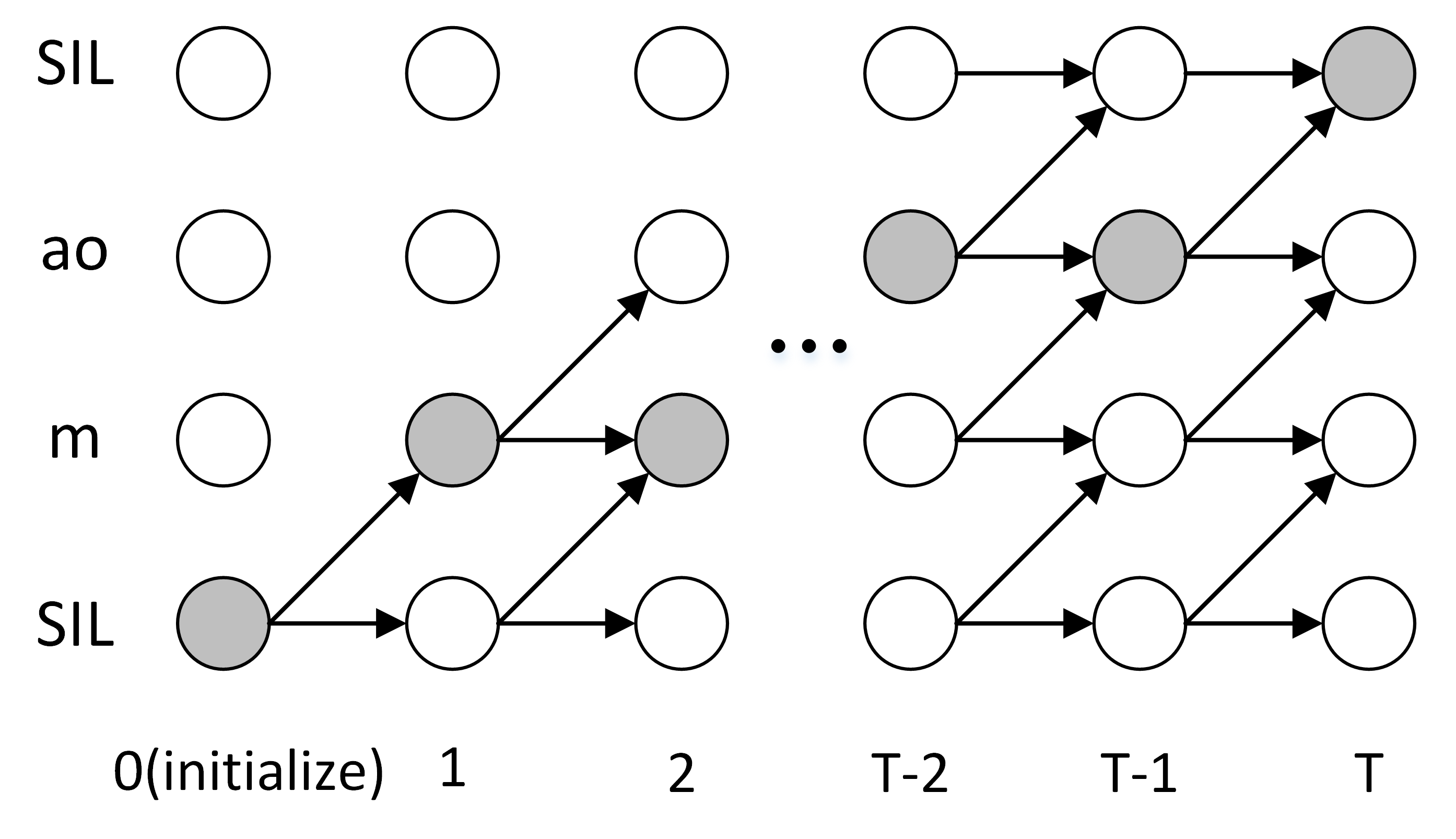}
\caption{Grey circles represent a possible alignment path. The alignment paths composed of arrows satisfy
$\{\pi_0, \pi_1,...,\pi_t\}\in\P$.}
\label{fig:illustration}
\end{figure}

Assuming $\pi_t$ at different decoder timesteps are conditionally independent given encoding results $\x$ and query $\q_t$, we can write the probability of an alignment path $\pi_{1:t}=\{\pi_{1},\dots,\pi_{t}\}$ as
\begin{equation}
p(\pi_{1:t}|\x,\q_{1:t})=\prod_{t'=1}^t{p(\pi_{t'}|\x,\q_{t'})}=\prod_{t'=1}^t{y_{t'}(\pi_{t'})}.
\label{eq6}
\end{equation}
We introduce a constant $\pi_0=1$ for initialization and the
probability of the alignment path $\{\pi_0, \pi_1,...,\pi_t\}$ can also be defined using Equation~(\ref{eq6}).
Let $\mathcal{P}$ denote the space of alignment paths
in which each path 
moves monotonically and continuously without skipping any encoder states.
Fig.~\ref{fig:illustration} gives an illustration of the alignment path when decoding acoustic features from an input phone sequence /SIL m ao SIL/ for speech synthesis.

Similar to the connectionist temporal classification (CTC) model \cite{graves2006connectionist},
a forward variable $\alpha_t(n)$ is defined here
to be the total probability of $\{\pi_0,\pi_1,...,\pi_t\}\in\P$ and $\pi_{t}=n$ as
\begin{equation}
\alpha_t(n) \overset{def}{=} \sum_{\substack{
 \pi_{0:t}\in\P \\
 \pi_{t}=n
}}{\prod_{t'=1}^t{y_{t'}(\pi_{t'})}}.
\end{equation}
Notice that $\alpha_t(n)$ can be calculated recursively from $\alpha_{t-1}(n)$ and $\alpha_{t-1}(n-1)$ as
\begin{equation}
\alpha_t(n)=(\alpha_{t-1}(n)+\alpha_{t-1}(n-1))y_t(n).
\end{equation}
Then we define
\begin{equation}
\hat{\alpha}_t(n)
 \overset{def}{=} {\left.\alpha_t(n) \middle/
 \sum_n{\alpha_t(n)}\right.}
 \end{equation}
 to make sure the sum of $\hat{\alpha}_t(n)$ for the $t$-th timestep to be $1$
and substitute $\hat{\alpha}_t(n)$  for $y_t(n)$ in Equation~(\ref{eq1}) to calculate the context vector as
\begin{equation}
\c_t=\sum_{n=1}^N{\hat{\alpha}_t(n)\x_{n}}.
\end{equation}
The complete forward attention method is described in Algorithm 1.
\begin{algorithm}[H]
\caption{Forward Attention}
\label{alg:forward attention}
\begin{algorithmic}
\Initialize{\strut$\hat{\alpha}_0(1) \gets 1$ \\ $\hat{\alpha}_0(n) \gets 0$, $n=2,...,N$}
\For{$t=1$ to $T$}
\State $y_t(n) \gets Attend(\x,\q_t)$
\State $\hat{\alpha}'_t(n) \gets (\hat{\alpha}_{t-1}(n)+\hat{\alpha}_{t-1}(n-1))y_t(n)$
\State $\hat{\alpha}_t(n) \gets \left.\hat{\alpha}'_t(n) \middle/ {\sum_{m=1}^N{\hat{\alpha}'_t(m)}}\right.$
\State $\c_t \gets \sum_{n=1}^N{\hat{\alpha}_t(n)\x_n}$
\EndFor
\end{algorithmic}
\end{algorithm}

\subsection{Forward Attention with Transition Agent}
\label{ssec:forward attention with TA}
A strategy of transition agent (TA) is further designed to help forward attention control the action of moving forward or staying during alignment flexibly.
Specifically, a transition agent DNN with one hidden layer and
sigmoid output activation unit is adopted to produce a scalar $u_t\in(0,1)$ for each decoder timestep.
$u_t$ can be considered as an indicator which describes the probability that the attended phone should move forward to the next one at the $t$-th decoder timestep.
$\c_t$, $\bm{o}_{t-1}$ and $\q_t$ are concatenated as the input of this DNN.
We simply integrate $u_t$ into the calculation of $\alpha_t(n)$ as shown in Algorithm 2.

\begin{algorithm}[H]
\caption{Forward Attention with Transition Agent}
\label{alg:forward attention ta}
\begin{algorithmic}
\Initialize{\strut$\hat{\alpha}_0(1) \gets 1$ \\
$\hat{\alpha}_0(n) \gets 0$, $n=2,...,N$ \\
$u_0 \gets 0.5$}
\For{$t=1$ to $T$}
\State $y_t(n) \gets Attend(\x,\q_t)$
\State $\hat{\alpha}'_t(n) \gets ((1-u_{t-1})\hat{\alpha}_{t-1}(n)+u_{t-1}\hat{\alpha}_{t-1}(n-1))y_t(n)$
\State $\hat{\alpha}_t(n) \gets \left.\hat{\alpha}'_t(n) \middle/ \sum_{m=1}^N{\hat{\alpha}'_t(m)}\right.$
\State $\c_t \gets \sum_{n=1}^N{\hat{\alpha}_t(n)\x_n}$
\State $u_t \gets DNN(\c_t,\bm{o}_{t-1},\q_t)$
\EndFor
\end{algorithmic}
\end{algorithm}

The method of forward attention with transition agent can also be explained from the point of view of
a product-of-experts model (PoE) \cite{hinton1999products, welling2007product}.
A PoE model combines a number of individual component models (the experts)
 by taking their product and normalizing the result.
 Each component in a product represents a soft constraint.
In our proposed forward attention with transition agent,
one expert $(1-u_{t-1})\hat{\alpha}_{t-1}(n)+u_{t-1}\hat{\alpha}_{t-1}(n-1)$
describes the constraint of monotonic alignment.
Another expert is the original attention probability given by $y_t(n)$.
The calculation of $\hat{\alpha}_t(n)$ is based on the product of these two experts.
Therefore, the alignment paths that violate the monotonic condition
are expected to have low probability.

Furthermore, the transition agent provides us an opportunity to control the speed of synthetic speech conveniently,
which is usually difficult for seq2seq acoustic modeling due to the lack of explicit duration models.
When we add positive or negative bias to the sigmoid output units of
the  transition agent DNN during generation,
the transition probability $u_t$ gets increased or decreased.
This can lead to a faster or slower movement of the attended phones, corresponding to a faster or slower speed of synthetic speech.

\section{Experiments}
\label{ssec:experiments}

\subsection{Experimental Conditions}
\label{ssec:experiments}

A Mandarin speech database recorded by a female professional speaker was used in our experiments.
The duration of the database was 19.8 hours, which contained 13334 utterances of 16kHz speech data. The database was divided into a training set and a test set, which had 12219 and 1115 utterances respectively.
We built seq2seq acoustic models based on the framework of Tacotron\cite{wang2017tacotron}.
The target acoustic features were log magnitude spectrogram extracted with Hamming windowing, 50 ms frame length, 12.5ms frame shift, and 2048-point Fourier transform. Griffin-Lim algorithm \cite{griffin1984signal} was used to synthesize waveform from the predicted spectrogram.
We extracted input features from phone sequences,
which were simply composed of the phone label (61-dimension one-hot vector) and tone label (5-dimension one-hot vector) for each phone.
These two vectors were first embedded into 224 and 32 dimensional descriptions respectively, and then passed to separate pre-nets.
The pre-nets for phone and tone information had the same width as their embedding dimension.
The outputs of both pre-nets were concatenated to form the input of the encoder. We employed the reduction trick with $r=2$ in all experiments.

Altogether 9 seq2seq acoustic models were built for comparison.\footnote{Audio samples available on \url{https://jxzhanggg.github.io/ForwardAttention}}
They were divided into 3 groups, which used the conventional attention method introduced in Section~\ref{sec:previous works} (baseline) , the proposed forward attention method (FA), and the forward attention with transition agent (FA-TA) respectively.
The 3 systems in each group adopted the windowing technique, the convolutional features, or none of them.
For the windowing technique, we set $w=2$.
For using convolutional features, we used $k=10$ and $l=5$ in our experiments.
We tried to train a system with location-based attention \cite{graves2013generating}.
However, the model failed to converge in our experiments.

We also built a LSTM-based system \cite{fan2014tts} for comparison.
41-dimension mel-cepstral coefficients (MCCs), and $F_0$ in log-scale were extracted every 5ms
using STRAIGHT\cite{kawahara1999restructuring}.
The LSTM acoustic model had 2 hidden layers and 512 units per layer.
The model inputs include 523 binary features for categorical linguistic contexts (e.g.
phones and tones identities, stress marks) and 9 numerical linguistic contexts (e.g. the number frames and position of current frame in a phone).
A separate DNN-based duration model was constructed to predict state durations at synthesis time.
The DNN had 3 hidden layers and 1024 units per layer, using 523-dimension binary linguistic contexts as input.


\subsection{Stability of Sequence-to-Sequence Feature Generation}
\label{ssec:robustness}
\begin{table}[tb]
    \centering
    \caption{Number of failed samples for the 9 evaluated seq2seq models, where ``Window" stands for using the windowing technique, ``Conv. Feats." stands for adding convolutional features and ``Plain" stands for using none of these two techniques.}\label{table:failed samples}
    \begin{tabular}{lccc}
    \toprule
    Model &Plain &Window &Conv. Feats.\\
    \midrule
    Baseline &54 &26 &7\\
    FA &5 &4 &\textbf{0}\\
    FA-TA &6 &3 &\textbf{0}\\
    \bottomrule
    \end{tabular}
\end{table}

We first evaluated the stability of acoustic feature generation using the 9 built seq2seq models with different attention mechanism.
120 utterances were randomly selected from the test set and synthesized using these systems.
The longest utterance had about 100 phones.
An experienced speech synthesis researcher was asked to listen to all these synthetic samples
and label the failed samples, i.e., the synthetic utterances with repeating phones, missing phones, or any kind of perceivable mistakes.
The results are summarized in Table~\ref{table:failed samples}.

\begin{figure}[tb]
  \centering
  \includegraphics[width=8.5cm]{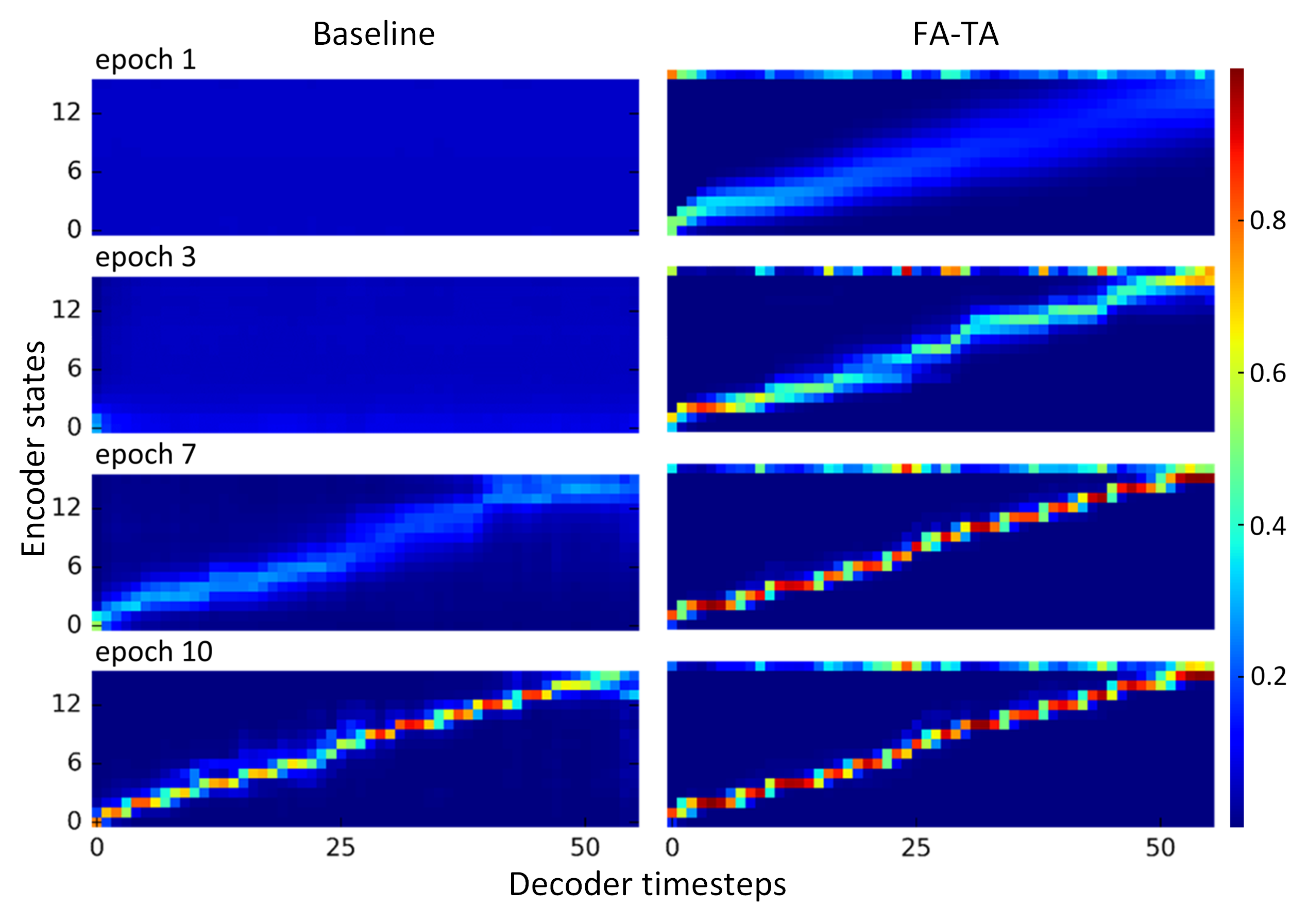}
\caption{Alignments of an utterance given by the baseline system and the FA-TA system after 1, 3, 7 and 10 epochs training.
The top row of each subgraph in the FA-TA column shows the transition probability $u$ predicted by the transition agent, and the rest rows show $\hat{\alpha}_t(n)$ in Algorithm ~\ref{alg:forward attention ta}.}
\label{fig:alignments}
\end{figure}

As we can see from this table, the baseline system with plain content-based attention
suffered from the mistakes made in the synthetic speech.
A close examination showed that this was caused by the inappropriate alignments
given by the attention probabilities.
Mistakes occurred when the alignment had aliasing, became disconnected, or got stuck at the same position.
By introducing the windowing technique or using convolutional features,
the performance of stability always got improved.
The two forward attention methods achieved better stability than the baseline attention method.
The best systems adopted forward attention (with or without transition agent) and convolutional features.
Moreover, we found that the forward attention systems converged much faster than the baseline systems.
Fig.~\ref{fig:alignments} shows  how the alignment changed in the plain baseline system and the plain FA-TA system after 1, 3, 7 and 10 epochs of model training.

\subsection{Naturalness of Synthetic Speech}
\label{ssec:naturalness}
\setlength{\tabcolsep}{5.2pt}
\begin{table}[tb]
  \centering
  \caption{Average preference scores(\%) on naturalness, where
  ``*" stands for using convolution features.
  ``N/P'' stands for no preference.
  $p$ denotes the $p$-value of a $t$-test between two systems.}\label{pref}
  \begin{tabular}{ccccccc}
  \toprule
  FA        &FA-TA          &FA-TA*     &Baseline   &LSTM       &N/P        &$p$ \\
  \midrule
  22.0      &\textbf{51.5}  &-          &-          &-          &26.5       &$<10^{-5}$ \\
  -         &\textbf{43.0}  &19.0       &-          &-          &38.0       &$<10^{-5}$\\
  -         &\textbf{43.0}  &-          &13.5       &-          &43.5       &$<10^{-5}$ \\
  -         &44.5           &-          &-          &37.5       &18.0       &0.275 \\

  \bottomrule
  \end{tabular}
\end{table}

Several groups of preferences were conducted to evaluate the naturalness of synthetic speech using different systems.
20 sentences which were correctly synthesized by all systems in the experiment of Section \ref{ssec:robustness} were adopted to generate the stimuli.
In each preference test, the utterances synthesized by two comparative systems were evaluated in random order by 10 native listeners using headphones.
The listeners were asked to judge which utterance in each pair had better naturalness or there was no preference.

We first compared the plain FA system, the plain FA-TA system, and the FA-TA system using convolutional features.
The results are shown in the first two rows of Table~\ref{pref}.
The results show the advantage of transition agent and the negative effect of adding convolutional features on the naturalness of synthetic speech. One possible reason is that convolutional features acted as a constrain of alignment and impaired the prosodic modeling capacity of attention mechanism.
Then, we conducted similar experiments to compare the FA-TA system with the plain baseline system and the conventional LSTM system.
The results shown in the last two rows of Table~\ref{pref} demonstrate that the FA-TA system outperformed the baseline and
achieved comparable results to the LSTM system.
We should notice that the LSTM system employed rich linguistic information as input while the FA-TA system only
used phone and tone labels for acoustic modeling.

\subsection{Speed Control Using Forward Attention with Transition Agent}
\label{ssec:speed}

\begin{figure}[tb]
  \centering
  \includegraphics[scale=0.6]{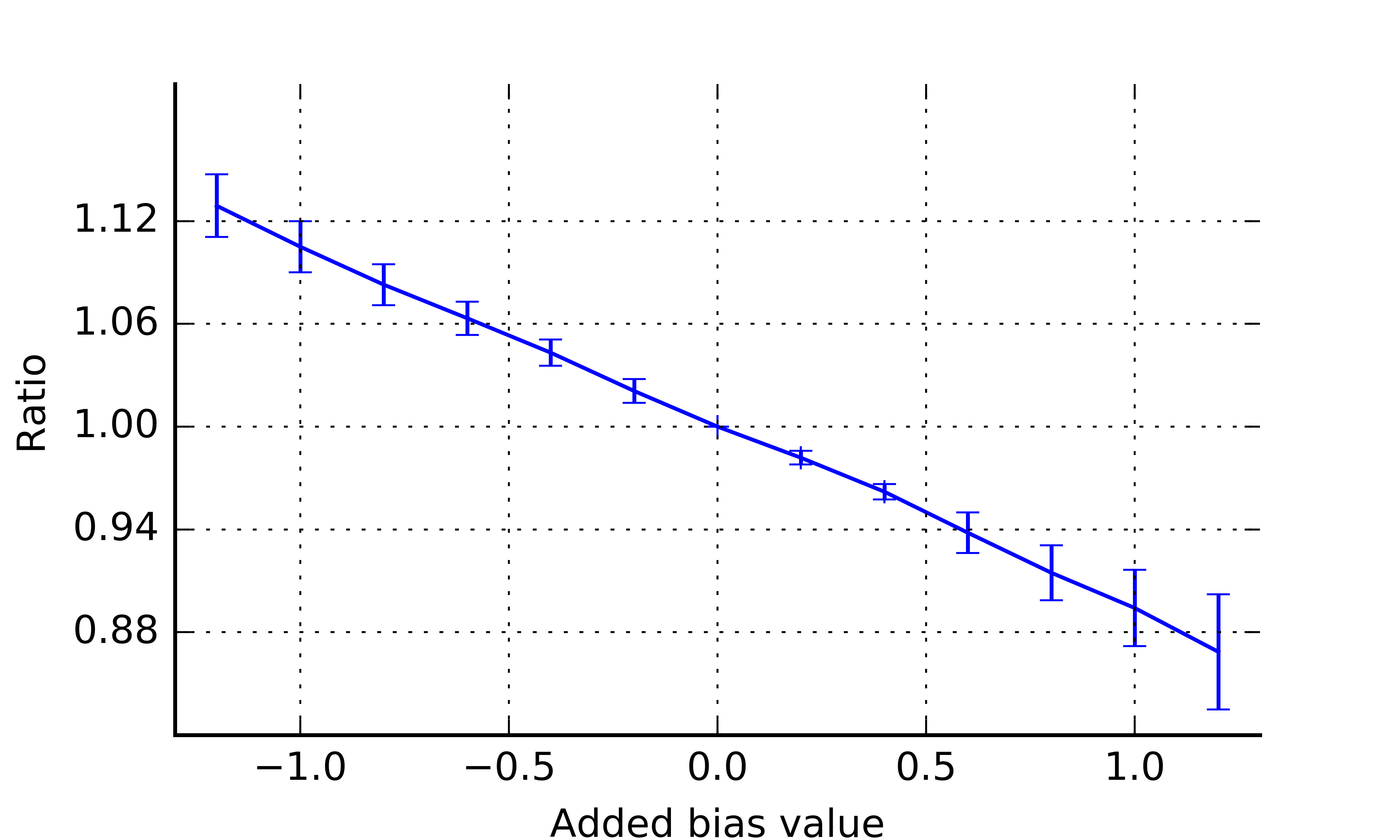}
\caption{Average ratios of sentence duration modification achieved by controlling the bias value in the FA-TA system. Error bars represent the standard deviations.}
\label{fig:speed}
\end{figure}

In the proposed forward attention with transition agent, as we adding positive or negative bias to the sigmoid output units of
the DNN transition agent during generation,
the transition probability gets increased or decreased.
This leads to a faster or slower of attention results.
An experiment was conducted using the plain FA-TA system to evaluate the effectiveness of speed control using this property.
We used the same test set of the 20 utterances in Section~\ref{ssec:naturalness}.
We increased or decreased the bias value from 0 with a step of 0.2, and synthesized all sentences in the test set.
We stopped once one of the generated samples had the problem of missing phones, repeating phones,
or making any perceivable mistakes.
Then we calculated the average ratios between the lengths of synthetic sentences using modified bias and the lengths of  synthetic sentences without bias modification.
Fig.~\ref{fig:speed} show the results in a range of bias modification where all samples were generated correctly.
From this figure, we can see that more than $10\%$ speed control can be achieved using the proposed forward attention with transition agent. Informal listening test showed that such modification did not degrade the naturalness of synthetic speech.

\section{Conclusions}
\label{sec:conclusions}
A forward attention method in the seq2seq acoustic modeling
for speech synthesis has been proposed.
Experimental results show that this method has the advantages of faster convergence during model training,
higher stability of acoustic feature generation, and feasibility of controlling the speed of synthetic speech.
This paper applies the proposed forward attention method to the speech synthesis task.
This method can also be modified and adapted to other tasks, such as speech recognition
and other seq2seq problems having the nature of monotony. Investigation on the performance of forward attention in these tasks will be apart of our future work.



\vfill\pagebreak


\bibliographystyle{IEEEbib}
\bibliography{refs}

\end{document}